\documentclass[letterpaper,journal]{IEEEtran}
\usepackage{amsmath,amsfonts,amssymb}
\usepackage{algorithm}
\usepackage{array}
\usepackage{textcomp}
\usepackage{stfloats}
\usepackage{url}
\usepackage{graphicx}
\usepackage{cite}
\usepackage{booktabs}
\usepackage[table]{xcolor}
\usepackage[hidelinks]{hyperref}
\newtheorem{assumption}{Assumption}
\newtheorem{proposition}{Proposition}
\newtheorem{corollary}{Corollary}

\definecolor{evicastblue}{RGB}{26,82,118}
\definecolor{evicastbg}{RGB}{226,239,248}
\definecolor{evicastgreen}{RGB}{38,138,84}
\newcommand{\strongresult}[1]{\cellcolor{evicastbg}\textcolor{evicastblue}{\textbf{#1}}}
\newsavebox{\evicasttablebox}
\newcommand{\fitwidth}[2]{%
  \sbox{\evicasttablebox}{#2}%
  \ifdim\wd\evicasttablebox>#1\relax
    \resizebox{#1}{!}{\usebox{\evicasttablebox}}%
  \else\usebox{\evicasttablebox}\fi}

\begin{document}
\bstctlcite{IEEEexample:BSTcontrol}
\title{EviCast: Forecasting Retrieval Gains Through Shared Response Calibration}
\author{%
Yibo~Wang,
Jisheng~Dang,
Bimei~Wang,
Jiongpei~Hou,
Jingwen~Zhao,
Wenjun~Ma,
Fei~Shen,
Juncheng~Li,
Meng~Wang,~\IEEEmembership{Fellow,~IEEE,}
Qi~Tian,~\IEEEmembership{Fellow,~IEEE,}
Bin~Hu,~\IEEEmembership{Fellow,~IEEE,}
and~Tat-Seng~Chua%
\thanks{Yibo Wang, Jisheng Dang, Bimei Wang, Jiongpei Hou, Jingwen Zhao, Wenjun Ma, and Bin Hu are with Lanzhou University, Lanzhou, China.}%
\thanks{Fei Shen is with Nanjing University of Science and Technology, Nanjing, China.}%
\thanks{Juncheng Li is with Zhejiang University, Hangzhou, China.}%
\thanks{Tat-Seng Chua is with the School of Computing, National University of Singapore, Singapore.}%
\thanks{Qi Tian is with Cloud and AI BU, Huawei, Shenzhen, Guangdong 518129, China (e-mail: tian.qi1@huawei.com).}
\thanks{Meng Wang is with the School of Computer Science and Information Engineering, Hefei University of Technology, Hefei 230009, China. (E-mail: eric.mengwang@gmail.com).}
\thanks{Corresponding authors: Bin Hu and Jisheng Dang.}%
}
\maketitle

\begin{abstract}
Retrieval-augmented generation (RAG) extends large language models with external evidence, yet relevant evidence does not ensure that a generator can use it. Indiscriminate retrieval can degrade task performance while adding retrieval and long-context inference costs. Existing methods neither determine before large-scale execution whether a generator, task distribution, and retriever configuration is worth deploying nor diagnose multiple candidate retrievers under a shared budget. We introduce EviCast, a framework that forecasts retrieval value through interventional evidence-state decomposition. EviCast measures a generator's response to controlled evidence states once and combines this shared profile with the offline evidence-state distribution of each candidate retriever, yielding the end-to-end net utility and its 95\% lower confidence bound without executing every RAG configuration. A state-refinement analysis characterizes when its binary and topology-aware instantiations are exact and when shared calibration is statistically more efficient than per-retriever calibration. In frozen evaluations, EviCast correctly decides all 14 HotpotQA retrieval configurations with a forecast error of 2.75 F1 points while reducing diagnostic generation calls by 80\%, and it attains the lowest forecast error among seven predictors in the low-budget regime. Its topology-aware instantiation, EviCast-T, captures the complementarity between terminal and bridge evidence, correctly deciding all 16 configurations on MuSiQue and HotpotQA and 23 of 24 after prospective transfer to a third dataset. These results show that retrieval deployment depends on both evidence availability and the generator's ability to convert evidence structure into stable net utility. Code is available at \url{https://github.com/YiboWang2002/EviCast}.
\end{abstract}


\section{Introduction}
\label{sec:introduction}

\IEEEPARstart{L}{arge} language models (LLMs) excel at knowledge-intensive question answering, content generation, and complex reasoning. However, their parametric knowledge is slow to update, incomplete, and difficult to trace to source material. Retrieval-augmented generation (RAG) introduces updatable and attributable external knowledge at inference time. It has therefore become a fundamental interface between LLMs and real-world knowledge bases \cite{lewis2020retrieval}.

\begin{figure}[!tb]
  \centering
  \includegraphics[width=\linewidth]{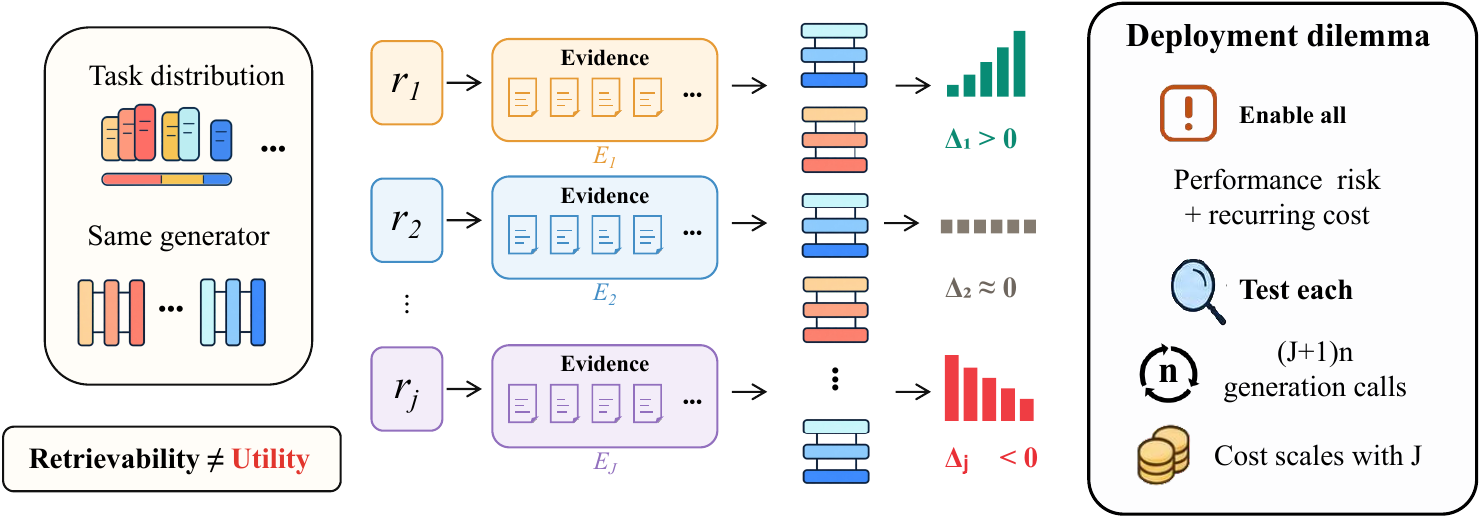}
  \caption{Retrieval utility and diagnosis under a fixed task distribution and generator.}
  \label{fig:motivation}
\end{figure}

RAG does not benefit every model or task. Its value depends on both retrieval relevance and the generator's ability to use the retrieved evidence. Irrelevant context can distract LLMs and induce incorrect answers \cite{shi2023distracted,amiraz2025distracting}. Indiscriminate retrieval may therefore harm performance while adding retrieval and long-context inference costs, whereas overly conservative retrieval may forfeit substantial gains from external evidence. Figure~\ref{fig:motivation} illustrates this deployment dilemma. We ask whether the effect of external evidence on a target model can be predicted before deployment and used to enable RAG only when beneficial. We further ask whether one diagnosis can serve multiple candidate retrieval schemes. These questions define retrieval necessity and its associated resource-efficiency problem.

Existing work studies retrieval decisions and utility at three levels. Forward-Looking Active Retrieval augmented generation (FLARE), Self-Reflective Retrieval-Augmented Generation (Self-RAG), and Adaptive-RAG trigger retrieval for an individual query or generation step \cite{jiang2023active,asai2024selfrag,jeong2024adaptive}. Retrieval Augmented Generation Assessment (RAGAS) and RAGChecker evaluate RAG systems after execution \cite{es2024ragas,ru2024ragchecker}. SePer, passage utility models, InfoGain-RAG, and retrieval performance prediction estimate the local utility of a query or document \cite{dai2025seper,perez2025uncertainty,wang2025infogain,tian2026predicting}. These methods do not determine before large-scale execution whether a complete generator, task distribution, and retriever configuration is worth deploying, nor do they diagnose multiple candidate retrievers under a shared generation budget. We formulate this problem as amortized retrieval necessity diagnosis at deployment time. Given a generator $g$, task distribution $\mathcal{D}$, and candidate retriever $r$, the objective is to predict the expected gain $\Delta(g,\mathcal{D},r)$ over the no-retrieval baseline and make a deployment decision.

Diagnosing retrieval necessity presents two technical challenges. First, the evidence representation must reflect how the generator uses evidence because RAG utility depends jointly on retrieved evidence and the generator's ability to exploit it. Relevance and correct-passage counts are insufficient. In multihop question answering, one top-2 result may contain a bridge passage and a distractor, while another contains a terminal answer passage and a distractor. Both contain one correct passage but may affect the final answer differently. Second, the generator response must transfer reliably across retrievers without executing every candidate RAG configuration, while accounting for uncertainty from small calibration samples. Directly comparing retrieval and no-retrieval performance for each retriever makes generation cost grow linearly with the number of candidates and yields variable small-sample gain estimates.

We address these challenges with EviCast, a retrieval necessity diagnosis framework based on interventional evidence-state calibration. EviCast decomposes candidate retriever utility into a generator response profile and a retriever-specific evidence-state distribution. Controlled evidence interventions estimate the response profile once on a small calibration set, while offline retriever outputs determine each state distribution. Their combination yields the expected net gain and its lower confidence bound (LCB). This decomposition reuses generator response calibration across retrievers, while topology-aware states capture complementary evidence roles in multihop reasoning.

Our frozen evaluation covers multiple RAG mechanisms, three multihop question-answering datasets, multimodal and text-only generators, and sparse, dense, and fusion retrievers. EviCast achieves 90\% decision agreement across 10 task configurations without falsely enabling any configuration that lacks a significant gain. On HotpotQA, 96 shared generations yield 14 correct decisions across 14 retrieval configurations, a mean absolute error (MAE) of 2.75 F1 points, and an 80\% reduction in diagnostic generation calls, and EviCast attains the lowest forecast error among seven predictors in the low-budget regime. Its topology-aware instantiation decides all 16 configurations on MuSiQue and HotpotQA correctly with an overall MAE of 2.81 points and 23 of 24 after prospective transfer to 2WikiMultiHopQA (2Wiki). Empirical checks further confirm each theoretical property. These results show that reusable calibration accurately forecasts retrieval value while substantially reducing the cost of diagnosing multiple configurations.

Our contributions are as follows:

\begin{itemize}
  \item We formulate deployment-level, amortized retrieval necessity diagnosis to predict gains over the no-retrieval baseline across candidate retrievers under a shared calibration budget. Decisions combine evidence-state separability and the utility LCB.

  \item We develop EviCast, a framework that calibrates generator responses through controlled evidence interventions, maps offline retriever outputs into evidence-state distributions, and combines both components into expected retrieval utility. We instantiate it with binary and topology-aware evidence states, where the latter models nonlinear complementarity between terminal and bridge evidence, and we prove when each instantiation is exact through a state-refinement analysis.

  \item We validate EviCast across multiple RAG mechanisms, three multihop question-answering datasets, two generators, and sparse, dense, and fusion retrievers. EviCast achieves 14/14 correct decisions on HotpotQA retrieval configurations, 16/16 on MuSiQue and HotpotQA topology configurations, and 23/24 after prospective transfer to a third dataset, attains the lowest forecast error among seven predictors in the low-budget regime, and reduces diagnostic generation calls by up to 80\%.
\end{itemize}

\section{Related Work}
\label{sec:related_work}

\subsection{Retrieval-Augmented Generation And Evidence Use}

Retrieval-augmented language models extend parametric knowledge with external memory. Early work established differentiable retrieval, dense passage retrieval, and joint retrieval and generation \cite{guu2020realm,karpukhin2020dpr,lewis2020retrieval}. Later systems fuse multiple passages, retrieve from large corpora, or augment frozen models with in-context evidence \cite{izacard2021fid,borgeaud2022retro,ram2023incontext}. Retrieval-Augmented Dual Instruction Tuning (RA-DIT) jointly adapts the generator and retriever, while RankRAG unifies context ranking with answer generation \cite{lin2024radit,yu2024rankrag}. Knowledge Intensive Language Tasks (KILT) provides a shared benchmark for knowledge-intensive tasks and provenance \cite{petroni2021kilt}.

External knowledge also supports visual question answering. Wu et al.~\cite{wu2018attributesknowledge} integrate visual attributes and external knowledge, while FVQA~\cite{wang2018fvqa} annotates questions with supporting facts. CRIC~\cite{gao2023cric} studies compositional reasoning over visual and commonsense evidence and evaluates grounding alongside answer correctness. More recently, ReT-2~\cite{caffagni2026ret2} develops universal multimodal retrieval and evaluates its use in retrieval-augmented visual question answering. These studies of knowledge integration, evidence grounding, and retrieval effectiveness complement EviCast's focus on forecasting the gain of a candidate retrieval configuration for a fixed generator.

External evidence is not uniformly beneficial. Irrelevant passages can distract generators, motivating evidence filtering and robustness training \cite{shi2023distracted,yoran2024robust,amiraz2025distracting}. Evidence position and document composition also alter model behavior \cite{liu2024lost,cuconasu2024noise}, while retrieval gains vary with the knowledge already stored by the model \cite{mallen2023trust}. These findings motivate EviCast to model retrieval supply and generator response separately rather than equating relevance with end-to-end utility.

\subsection{Adaptive Retrieval And Selective Augmentation}

Adaptive methods trigger retrieval from generation confidence, model self-knowledge, reflection, question complexity, or real-time information needs \cite{jiang2023active,wang2023skr,asai2024selfrag,jeong2024adaptive,su2024dragin}. Selective augmentation instead compresses or filters retrieved context before generation \cite{xu2024recomp,wang2023filco}. Unlike these query-level or step-level methods, EviCast forecasts configuration-level deployment utility using one generator calibration shared across candidate retrievers.

Dynamic neural networks provide a broader perspective on allocating computation according to the input~\cite{han2022dynamic}. For example, Glance and Focus Networks~\cite{huang2023glance} first inspect a coarse global view, then process selected image regions and stop when predictions are sufficiently confident. Such inference-time adaptation motivates efficient evidence use. EviCast makes a different resource decision before deployment: whether a retrieval configuration yields a reliable expected gain over the target task distribution.

\subsection{RAG Evaluation, Benchmarks, And Utility Estimation}

RAGAS, the Automated RAG Evaluation System (ARES), and RAGChecker evaluate executed pipelines through context relevance, faithfulness, answer quality, and component-level diagnostics \cite{es2024ragas,saadfalcon2024ares,ru2024ragchecker}. The Retrieval-Augmented Generation Benchmark (RGB), Automatic LLMs' Citation Evaluation (ALCE), and RAGTruth assess robustness, information integration, citation quality, and hallucination \cite{chen2024rgb,gao2023alce,niu2024ragtruth}. These tools characterize RAG behavior after execution. Utility modeling is closest to our objective. Prior work estimates the value of a query or document through semantic perplexity reduction, learned passage utility, confidence change, and retrieval-quality features \cite{dai2025seper,perez2025uncertainty,wang2025infogain,tian2026predicting}. EviCast instead decomposes configuration-level net gain into interventional generator responses and retriever-specific evidence-state distributions, enabling amortized deployment diagnosis.

Confidence estimation also highlights the need to match reliability measures to the intended decision. Zhu et al.~\cite{zhu2024confidence} show that improving confidence calibration need not improve failure prediction. In EviCast, calibration denotes measuring task-score responses to controlled evidence states; the deployment criterion concerns a confidence bound on expected retrieval gain rather than the confidence of an individual answer.

\subsection{Multimodal Evidence And Reliable Fusion}

Multimodal learning studies how heterogeneous sources can be represented, aligned, and fused~\cite{baltrusaitis2019multimodal}, including cross-modal interactions implemented with Transformers~\cite{xu2023multimodaltransformers}. Video models aggregate temporal evidence through spatio-temporal memory~\cite{dang2023imsfr,dang2024stcm,dang2024tspsm,dang2025asm,dang2026vdn,dang2026sstp}, decoupled segmentation tokens~\cite{dang2025desa2va}, and preference-optimized captioning~\cite{dang2025synpo}, and omni-modal models ground affective reasoning in audiovisual and paralinguistic cues~\cite{wang2026affectomni,wang2026dualaffect,wang2026readsteer}. Trusted Multi-View Classification~\cite{han2023trusted} models view-specific uncertainty through dynamic evidential fusion, underscoring the source reliability that also matters in our acoustic, visual, and language retrieval settings. EviCast evaluates a fixed generator and evidence interface: it measures how evidence conditions affect task utility and uses those responses to forecast retrieval gains, without learning a new fusion architecture.

\section{Method}
\label{sec:method}

\begin{figure*}[!tb]
  \centering
  \IfFileExists{figures/evicast_framework_recreated_editable.pdf}{
    \includegraphics[width=\linewidth]{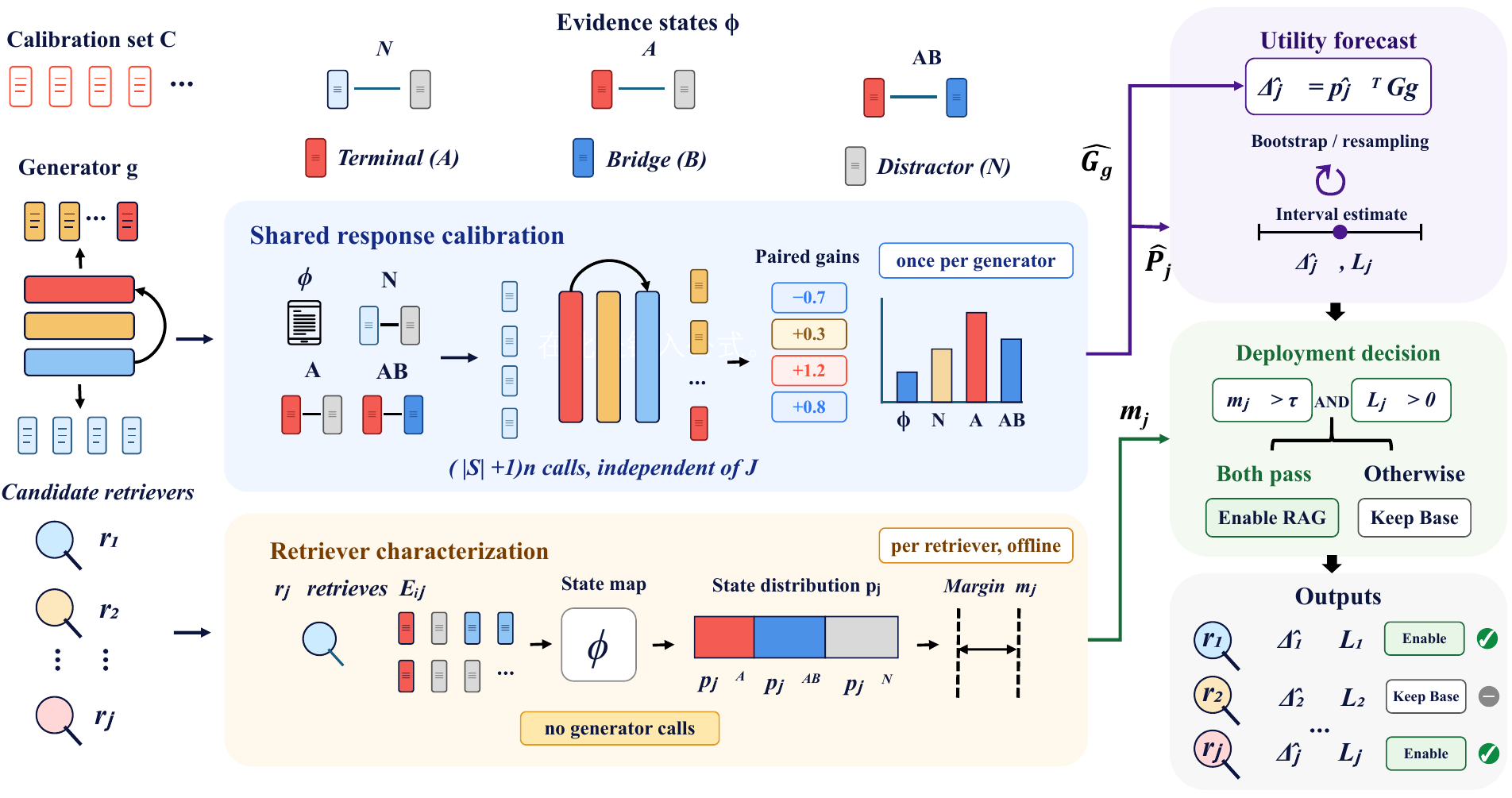}
  }{
    \fbox{\rule{0pt}{1.90in}\rule{0.95\linewidth}{0pt}}
  }
  \caption{EviCast framework for reusable retrieval utility forecasting and deployment.}
  \label{fig:evicast}
\end{figure*}

EviCast forecasts end-to-end RAG utility before deployment. It measures a generator's response to controlled evidence states, then estimates how often each candidate retriever produces these states in offline results. Combining the shared response profile with each retriever distribution yields an expected net gain, a confidence interval (CI), and a deployment decision. This decomposition avoids complete RAG calibration for every retriever. Figure~\ref{fig:evicast} shows the full data flow.

EviCast is a framework rather than a single estimator. It is specified by an evidence-state map, a set of controlled interventions that realize each state, and the shared estimation and decision rule of Sections~\ref{sec:response_calibration} to~\ref{sec:utility_decision}. We develop two instantiations that share every component except the state map. The binary instantiation, EviCast-B, labels each retrieved unit as supportive or misleading and suits mutually substitutable evidence such as retrieved cases, neighbor labels, or single supporting passages. The topology-aware instantiation, EviCast-T, distinguishes terminal and bridge roles and suits compositional evidence whose units contribute complementary information. Section~\ref{sec:theory} characterizes when each instantiation is exact.

\subsection{Problem Formulation}
\label{sec:problem}

We define a fixed deployment environment as $\mathcal{E}=(g,\mathcal{D},\pi,\mathcal{K},k)$, where $g$ is the generator, $\mathcal{D}$ the target task distribution, $\pi$ the prompt template, $\mathcal{K}$ the retrieval corpus, and $k$ the number of returned passages. Given candidate retrievers $\mathcal{R}=\{r_1,\ldots,r_J\}$, retriever $r_j$ returns $E_j(x)=r_j(x;\mathcal{K},k)$. Higher values of the task metric $M(\hat y,y)$ indicate better predictions.

The true deployment utility of $r_j$ relative to no retrieval is:
\begin{equation}
  \begin{aligned}
  \Delta_j=\mathbb{E}_{(x,y)\sim\mathcal{D}}\bigl[
      &M\!\left(g_{\pi}(x,E_j(x)),y\right)\\
      &-M\!\left(g_{\pi}(x,\emptyset),y\right)\bigr].
  \end{aligned}
  \label{eq:true_utility}
\end{equation}
The ideal deployment action is:
\begin{equation}
  a_j^{*}=\mathbf{1}[\Delta_j>0],
  \label{eq:ideal_decision}
\end{equation}
where one enables retrieval and zero retains no-retrieval inference.

Directly computing $\Delta_j$ requires separate RAG generations for every retriever. EviCast instead uses a labeled calibration set $\mathcal{C}=\{(x_i,y_i,\mathcal{P}_i,\lambda_i)\}_{i=1}^{n}$ disjoint from the final test set. Here $\mathcal{P}_i$ contains candidate evidence, and $\lambda_i$ contains relevance or role annotations used only for calibration. EviCast returns $\widehat\Delta_j$, its CI, and $\widehat a_j$. The generator, task distribution, prompt, corpus, retrieval depth, and evidence format remain fixed during diagnosis, while only the offline rankings vary. The objective is to recover the sign and magnitude of $\Delta_j$ with generation cost independent of $J$.

\subsection{Binary And Topology-Aware Evidence States}
\label{sec:evidence_states}

EviCast uses an \emph{evidence state} as its central intermediate variable rather than a relevance score or correct-passage count. For query $x$ and evidence set $E$, the state mapping is:
\begin{equation}
  s=\phi(x,E;\lambda)\in\mathcal{S},
  \label{eq:state_mapping}
\end{equation}
where $\mathcal{S}$ is a mutually exclusive and exhaustive state space. Each state should preserve the evidence properties that determine generator utility. Within task stratum $h$, the response should be approximately invariant to the retriever that produced the evidence. Let $\delta(x,E)=M(g_{\pi}(x,E),y)-M(g_{\pi}(x,\emptyset),y)$. The desired response-invariance condition is:
\begin{equation}
  \mathbb{E}[\delta(x,E)\mid h,s,r_j]
  \approx
  \mathbb{E}[\delta(x,E)\mid h,s].
  \label{eq:state_sufficiency}
\end{equation}
Retrievers may change state probabilities but should not systematically change the response within a state. This condition allows one generator response calibration to be reused across retrievers.

\noindent\textbf{Binary Instantiation.}\enspace EviCast-B labels each of the $k$ retrieved units as supportive or misleading. It intervenes only on the two endpoints, an all-supportive set and an all-misleading set, whose responses in stratum $h$ are $G_{h,+}$ and $G_{h,-}$. With $\rho_{j,h}$ denoting the mean fraction of supportive units returned by retriever $r_j$, its forecast is:
\begin{equation}
  \widehat\Delta^{\mathrm{B}}_j=\sum_{h\in\mathcal{H}}\widehat w_h\bigl[\widehat\rho_{j,h}\widehat G_{h,+}+(1-\widehat\rho_{j,h})\widehat G_{h,-}\bigr].
  \label{eq:binary_forecast}
\end{equation}
This is the general forecast of Section~\ref{sec:utility_decision} with the unit-level state space $\{+,-\}$, and it requires only $3n$ generator calls.

\noindent\textbf{Topology-Aware Instantiation.}\enspace Multihop question answering requires evidence with distinct, complementary roles. Let $e^{A}$ denote the terminal passage that directly contains answer information, and let $\mathcal{B}$ denote bridge passages that identify an intermediate entity or connect the reasoning chain. For a fixed top-2 evidence set $E$, the topology-aware state mapping is defined as follows:
\begin{equation}
  \phi(E)=
  \begin{cases}
    N,  & e^{A}\notin E,\\
    A,  & e^{A}\in E,\ E\cap\mathcal{B}=\emptyset,\\
    AB, & e^{A}\in E,\ E\cap\mathcal{B}\neq\emptyset.
  \end{cases}
  \label{eq:topology_state}
\end{equation}
State $N$ lacks terminal evidence, state $A$ contains only terminal evidence, and state $AB$ contains terminal evidence with at least one bridge passage. This topology distinguishes a bridge passage paired with a distractor from a terminal passage paired with a distractor, although both sets contain one annotated supporting passage.

For each calibration example, EviCast constructs length-matched interventions for all three states. The $N$ intervention excludes terminal evidence, prioritizes a bridge passage, and fills the remaining position with a distractor. The $A$ intervention contains terminal evidence and one distractor, while the $AB$ intervention contains terminal and bridge evidence. A prespecified anchor retriever selects and orders candidates after their roles are fixed. Evidence states, task strata, and ordering rules are frozen before calibration generation.

\subsection{Reusable Interventional Response Calibration}
\label{sec:response_calibration}

Observed retrieval alone cannot distinguish missing evidence from a generator's failure to use available evidence. EviCast therefore intervenes on evidence states to measure the generator response separately. For item $i$, the no-retrieval score, controlled score, and paired gain are:
\begin{equation}
  \begin{aligned}
    b_i&=M(g_{\pi}(x_i,\emptyset),y_i),\\
    q_{i,s}&=M(g_{\pi}(x_i,\widetilde E_{i,s}),y_i),\\
    d_{i,s}&=q_{i,s}-b_i,
  \end{aligned}
  \label{eq:item_response}
\end{equation}
where $\widetilde E_{i,s}$ is the controlled evidence for state $s$. Pairing reduces question-difficulty variation.

Let $\mathcal{C}_h=\{i:h_i=h\}$ and $n_h=|\mathcal{C}_h|$. The response of generator $g$ in stratum $h$ and state $s$ is:
\begin{equation}
  \widehat G_{g,h,s}
  =\frac{1}{n_h}\sum_{i\in\mathcal{C}_h}d_{i,s}.
  \label{eq:response_profile}
\end{equation}
All retrievers share the baseline and controlled generations. Thus, $\widehat G$ is computed once for a fixed generator, prompt, and evidence format. Changing retrievers requires no new response generation.

\subsection{Retriever Evidence-State Characterization}
\label{sec:retriever_characterization}

Each candidate retriever runs offline on the shared calibration set. Let $E_{ij}=r_j(x_i;\mathcal{K},k)$ and $s_{ij}=\phi(x_i,E_{ij};\lambda_i)$. The state probability for retriever $r_j$ within stratum $h$ is:
\begin{equation}
  \widehat p_{j,h,s}
  =\frac{1}{n_h}\sum_{i\in\mathcal{C}_h}\mathbf{1}[s_{ij}=s].
  \label{eq:state_distribution}
\end{equation}
This mapping places heterogeneous retrievers in a common state space and combines them with the shared response profile without additional generator calls.

EviCast also tests whether a retrieval representation separates task-relevant evidence from distractors. Let $A_j$ be a separability statistic and $A_0$ its uninformative baseline. The evidence separability margin is defined as follows:
\begin{equation}
  m_j=A_j-A_0.
  \label{eq:representation_margin}
\end{equation}
For ranked question answering, $A_j$ is the pairwise area under the receiver operating characteristic curve (AUC) for ranking supporting passages above distractors, and $A_0=0.5$. For case retrieval, $A_j$ is held-out prediction performance based on retrieval keys, and $A_0$ is the majority baseline. This excess statistic provides a common separability gate across retrieval mechanisms.

\subsection{Utility Forecast And Deployment Decision}
\label{sec:utility_decision}

The response profile quantifies the value of each evidence state, while the retriever distribution gives its frequency. The law of total expectation yields:
\begin{equation}
  \begin{aligned}
  \widehat{\Delta}_j
  &=\sum_{h\in\mathcal{H}}\widehat w_h
   \sum_{s\in\mathcal{S}}
   \widehat p_{j,h,s}\widehat G_{g,h,s},\\
  \widehat w_h&=\frac{n_h}{n}.
  \end{aligned}
  \label{eq:utility_forecast}
\end{equation}
The utility forecast weights the generator response to each state by the probability that the retriever supplies that state. Under Eq.~\eqref{eq:state_sufficiency}, a new retriever requires only a new estimate of $\widehat p$, not $\widehat G$.

Both terms are uncertain when estimated from a small calibration set. EviCast propagates their joint uncertainty through a paired stratified bootstrap. Each replicate samples calibration items with replacement within $h$ and recomputes the response profile, state probabilities, and utility. We use $B=10{,}000$ replicates. The utility LCB is:
\begin{equation}
  L_j=\operatorname{Quantile}_{0.025}
  \left(\left\{\widehat{\Delta}^{*(b)}_j\right\}_{b=1}^{B}\right).
  \label{eq:utility_lcb}
\end{equation}
The final deployment rule combining evidence separability with utility confidence is:
\begin{equation}
  \widehat a_j
  =\mathbf{1}[m_j>\tau]\mathbf{1}[L_j>0].
  \label{eq:deployment_rule}
\end{equation}
We fix $\tau=0.05$ before evaluation. Retrieval is enabled only when the representation accesses task-relevant evidence and the expected gain has a positive 95\% LCB. Forecast $\widehat\Delta_j$ supports retriever ranking and resource planning.

\subsection{Diagnostic Algorithm And Amortized Cost}
\label{sec:algorithm}

The diagnostic procedure, summarized in Appendix Algorithm~S1, calibrates the shared generator response before evaluating any number of retrievers. Retrieved evidence is used only to map outputs to states and estimate their distribution. It is not provided to the generator during response calibration.

With $n$ calibration items, $|\mathcal{S}|$ evidence states, and $J$ candidate retrievers, the diagnostic generation costs are:
\begin{equation}
  \begin{aligned}
  C_{\mathrm{EviCast}}&=(|\mathcal{S}|+1)n,\\
  C_{\mathrm{Direct}}&=(J+1)n.
  \end{aligned}
  \label{eq:diagnostic_cost}
\end{equation}

For EviCast-T with top-2 retrieval, $|\mathcal{S}|=3$, so EviCast uses $4n$ generator calls for a fixed generator regardless of $J$. Adding a retriever requires offline retrieval and state counting but no additional generation. The savings of shared calibration therefore grow linearly with the number of candidate retrievers. Appendix~\mbox{A-A} provides the notation summary and additional intervention details.

\subsection{Theoretical Properties}
\label{sec:theory}

The decomposition in Eq.~\eqref{eq:utility_forecast} is exact under the following condition, which sharpens Eq.~\eqref{eq:state_sufficiency}.

\begin{assumption}[State Sufficiency]\label{ass:sufficiency}
For every stratum $h$, state $s$, and retriever $r_j$ with $p_{j,h,s}>0$, the retrieved and controlled responses coincide:
\begin{equation*}
  \begin{aligned}
  &\mathbb{E}\bigl[\delta(x,E_j(x))\mid h,\ \phi(x,E_j(x))=s\bigr]\\
  &\qquad=\mathbb{E}\bigl[\delta(x,\widetilde E_s(x))\mid h\bigr]=:G_{h,s}.
  \end{aligned}
\end{equation*}
\end{assumption}

Under Assumption~\ref{ass:sufficiency}, $\Delta_j=\sum_h w_h\sum_s p_{j,h,s}G_{h,s}$ for every retriever, and the plug-in forecast is consistent (Appendix Proposition~S1). The remaining design question is how coarse a state map may be. Proposition~\ref{prop:refinement} answers it.

\begin{proposition}[State Refinement]\label{prop:refinement}
Let $\phi'$ refine $\phi$, so that each state $s$ of $\phi$ is a union of states $f\in F(s)$ of $\phi'$, and let Assumption~\ref{ass:sufficiency} hold for $\phi'$ with responses $G_{h,f}$. Suppose the intervention for $s$ realizes the composition $\kappa_h(\cdot\mid s)$ over $F(s)$, whereas retriever $r_j$ produces $p_{j,h}(\cdot\mid s)$. Then the population forecast $\widetilde\Delta^{\phi}_j$ under $\phi$ satisfies:
\begin{equation}
  \begin{aligned}
  \widetilde\Delta^{\phi}_j-\Delta_j
  =\sum_{h}w_h\sum_{s}p_{j,h,s}\!\!\sum_{f\in F(s)}\!\!
  \bigl(&\kappa_h(f\mid s)\\
  -\,&p_{j,h}(f\mid s)\bigr)G_{h,f},
  \end{aligned}
  \label{eq:refinement_bias}
\end{equation}
and hence $|\widetilde\Delta^{\phi}_j-\Delta_j|\le\sum_h w_h\sum_s p_{j,h,s}\,\mathrm{TV}_{j,h,s}\,\Gamma_{h,s}$, where $\mathrm{TV}_{j,h,s}$ is the total-variation distance between $\kappa_h(\cdot\mid s)$ and $p_{j,h}(\cdot\mid s)$, and $\Gamma_{h,s}=\max_{f\in F(s)}G_{h,f}-\min_{f\in F(s)}G_{h,f}$.
\end{proposition}

A coarse state is therefore exact whenever the responses it merges coincide or every retriever reproduces the composition realized by the intervention. Merging terminal-only and bridge-only evidence violates both conditions, because their responses differ by 23 to 49 F1 points (Appendix Table~S6) and retrievers return them with different frequencies. The binary instantiation is the coarsest case and admits an exact characterization.

\begin{corollary}[Exactness Of EviCast-B]\label{cor:binary}
For top-2 retrieval, let Assumption~\ref{ass:sufficiency} hold for the number of supportive units $c\in\{0,1,2\}$ with responses $G_{h,c}$, and let $\gamma_h=G_{h,1}-\tfrac12(G_{h,0}+G_{h,2})$ denote the linearity gap. Then $\widetilde\Delta^{\mathrm{B}}_j-\Delta_j=-\sum_h w_h\,p_{j,h,1}\,\gamma_h$. EviCast-B is thus exact for every retriever whenever $\gamma_h=0$, and $\gamma_h$ is measurable before deployment with one additional partial-evidence intervention.
\end{corollary}

Shared calibration is also statistically efficient. Appendix Proposition~S2 shows that, under a common budget of generator calls, EviCast attains lower asymptotic variance than per-retriever Direct calibration whenever $J>J^\star_j=(|\mathcal{S}|+1)\nu_j^2/\sigma_j^2-1$, where $\nu_j^2$ and $\sigma_j^2$ are the influence-function variances of the two estimators. Section~\ref{sec:theory_check} verifies these properties empirically, and Appendix~E gives all proofs.
\section{Experiments}
\label{sec:experiments}

We evaluate EviCast along six dimensions. First, we assess retrieval necessity decisions across tasks and RAG paradigms. The cross-task settings cover Retrieval-Augmented Emotion Reasoning (RAER), Retrieval Augment for Missing Modality Multimodal Emotion Recognition (RAMER), and k-nearest-neighbor (kNN) label evidence on Explainable Multimodal Emotion Recognition (EMER), the Multimodal EmotionLines Dataset (MELD), and the Ryerson Audio-Visual Database of Emotional Speech and Song (RAVDESS). Here A, V, and L denote acoustic, visual, and language modalities. Second, we test whether response calibration transfers across candidate retrievers and reduces diagnostic cost. Third, we compare its utility forecasts with recent performance predictors under matched generation budgets. Fourth, we evaluate generalization across datasets, generators, and model scales. Fifth, state-representation ablations and controlled-response analyses examine the role of evidence topology. Finally, we verify the theoretical properties of Section~\ref{sec:theory} empirically. All experiments use disjoint calibration and test data, and every deployment decision is frozen before observing the final test results.

\begin{table*}[!tb]
  \centering
  \setlength{\tabcolsep}{4pt}
  \footnotesize
  \caption{EviCast retrieval decisions and independent test gains across 10 task configurations.}
  \label{tab:cross_task_gate}
  \fitwidth{\linewidth}{
  \begin{tabular}{@{}ccccccc@{}}
    \toprule
    Mechanism & Dataset / setting & Separability margin (pp) & Utility LCB (pp) & Decision & Test gain [95\% CI] & Agree \\
    \midrule
    RAER & EMER & +5.56 & +0.00 & Disable & $-0.25~[-0.76,0.00]$ & $\textcolor{evicastgreen}{\checkmark}$ \\
    RAER & MELD & $-12.05$ & +0.00 & Disable & $0.00~[0.00,0.00]$ & $\textcolor{evicastgreen}{\checkmark}$ \\
    kNN & RAVDESS & +62.50 & +18.46 & \strongresult{Enable} & \strongresult{$+27.50~[+19.50,+35.50]$} & $\textcolor{evicastgreen}{\checkmark}$ \\
    RAMER & EMER / A & +0.00 & $-12.18$ & Disable & $+3.69~[-2.01,+9.53]$ & $\textcolor{evicastgreen}{\checkmark}$ \\
    RAMER & EMER / V & +2.78 & $-7.34$ & Disable & $+1.32~[-2.36,+5.01]$ & $\textcolor{evicastgreen}{\checkmark}$ \\
    RAMER & EMER / L & +48.61 & $-3.80$ & Disable & $+6.71~[+1.54,+11.60]$ & $\times$ \\
    RAMER & EMER / AV & +2.78 & +0.41 & Disable & $-0.13~[-2.86,+2.57]$ & $\textcolor{evicastgreen}{\checkmark}$ \\
    RAMER & EMER / AL & +44.44 & $-4.34$ & Disable & $-3.37~[-9.55,+2.74]$ & $\textcolor{evicastgreen}{\checkmark}$ \\
    RAMER & EMER / VL & +16.67 & $-8.73$ & Disable & $-2.85~[-5.94,+0.16]$ & $\textcolor{evicastgreen}{\checkmark}$ \\
    Dense RAG & HotpotQA / BGE-M3 & +34.38 & +22.49 & \strongresult{Enable} & \strongresult{$+41.15~[+34.96,+47.34]$} & $\textcolor{evicastgreen}{\checkmark}$ \\
    \midrule
    \multicolumn{5}{c}{Overall} & \multicolumn{2}{c}{\cellcolor{evicastbg}\textcolor{evicastblue}{\textbf{9/10 agreement, 0 false enables}}} \\
    \bottomrule
  \end{tabular}}
\end{table*}

\subsection{Evaluation Protocol}
\label{sec:experimental_setup}

\noindent\textbf{Scope And Protocol.}\enspace The evaluation comprises 10 cross-task units spanning case-based emotion reasoning, missing-modality augmentation, label evidence, and dense passage RAG (Dense RAG) \cite{wen2025raer,fan2024ramer,ma2024emotion2vec,lewis2020retrieval}. We evaluate shared calibration across 14 HotpotQA retrieval configurations and topology-aware diagnosis on MuSiQue and HotpotQA with two generators and four dense, sparse, and fusion retrievers, followed by a prospective transfer to 2Wiki. A separately frozen transfer evaluates Qwen2.5 text-only generators at 3B and 7B scales. Every decision is fixed before independent testing, and calibration and test examples are disjoint. Within each comparison, all methods use the same queries, prompts, retrieval depth, evidence format, and answer parser. Appendices~A and~H provide complete splits, evidence-role rules, inference settings, and external-method adaptations.

\noindent\textbf{Evaluation.}\enspace We use mean example-level F1 for multi-label emotion recognition, accuracy for single-label classification, and normalized token F1 for open-ended question answering. We evaluate utility forecasts by their MAE against independent test gains and by Pearson and Spearman correlations. Gains, separability margins, utility bounds, and forecast errors are absolute differences reported in percentage points (pp) of the task metric, whereas percentages denote relative changes. A configuration has a significant positive gain only when the 95\% LCB of its paired test utility exceeds zero. We report agreement with this criterion, false enabling decisions, and diagnostic generation calls. Call counts exclude final test inference common to all methods. Appendix~A details baselines, state ablations, bootstrapping, and metrics.

\noindent\textbf{Models And Retrieval.}\enspace The multihop question-answering experiments use Qwen2.5-Omni-7B thinker (Qwen) and Phi-4-multimodal-instruct (Phi-4) in text-only mode. Both use deterministic greedy decoding, a batch size of 4, and at most 64 generated tokens. The four base retrievers return the top two full passages. BGE-M3 encodes the question and each title concatenated with its passage, then ranks candidates by cosine similarity. BM25 uses $k_1=1.5$ and $b=0.75$. Term frequency inverse document frequency (TF-IDF) uses cosine similarity over lowercased word unigrams and bigrams. Reciprocal Rank Fusion (RRF) combines BGE-M3 and BM25 rankings with $c=60$. Ties follow the original candidate order. Title and lead-sentence variants alter only the ranking text. The generator always receives the full passage.

\noindent\textbf{Calibration And Data Isolation.}\enspace The HotpotQA dense-RAG diagnosis and the 14-configuration reuse study share 32 calibration and 256 test queries. The topology-aware studies on MuSiQue and HotpotQA use 128 calibration and 512 test queries each, and the topology-aware HotpotQA split shares no example IDs with the reuse split. Stratification follows question hops on MuSiQue and question type and difficulty on HotpotQA. Prospective 2Wiki transfer and each text-only scale study also use 128 calibration and 512 disjoint test queries. All decisions are frozen before final evaluation; neither test answers nor test retrieval states are fed back into thresholds. Appendix~A gives the complete split counts, and Appendix~F records the isolation checks.

\subsection{Retrieval Necessity Across Tasks And RAG Paradigms}
\label{sec:cross_task}

Table~\ref{tab:cross_task_gate} reports the frozen EviCast decisions for 10 task configurations. EviCast agrees with the independent test conclusion in 9 of 10 cases, yielding 90\% decision accuracy with no false enabling decisions. It correctly enables kNN label evidence on RAVDESS and Dense RAG on HotpotQA, which yield test gains of 27.50 and 41.15 points, respectively. The remaining eight configurations are disabled, which avoids every configuration without a stable positive gain. These results span different objectives, evidence forms, and injection mechanisms, showing that EviCast diagnoses the end-to-end gain from external evidence rather than a local property of one RAG implementation.

\begin{figure}[!tb]
  \centering
  \includegraphics[width=\linewidth]{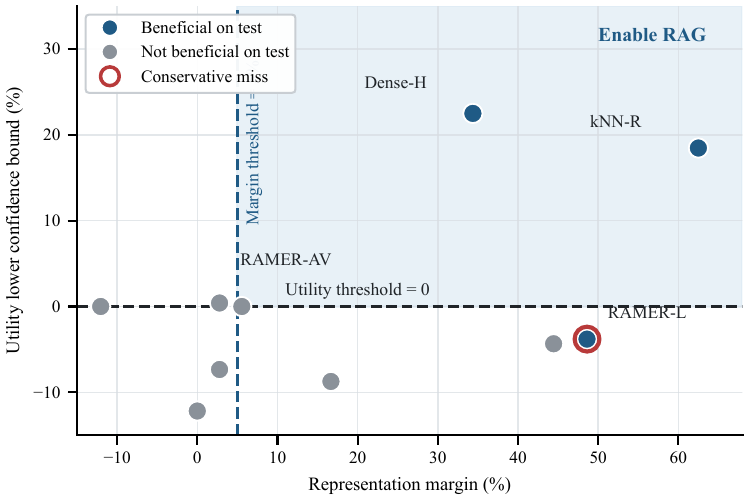}
  \caption{Two-gate decision plane for 10 cross-task configurations.}
  \label{fig:decision_plane}
\end{figure}

\begin{figure*}[!tb]
  \centering
  \includegraphics[width=\linewidth]{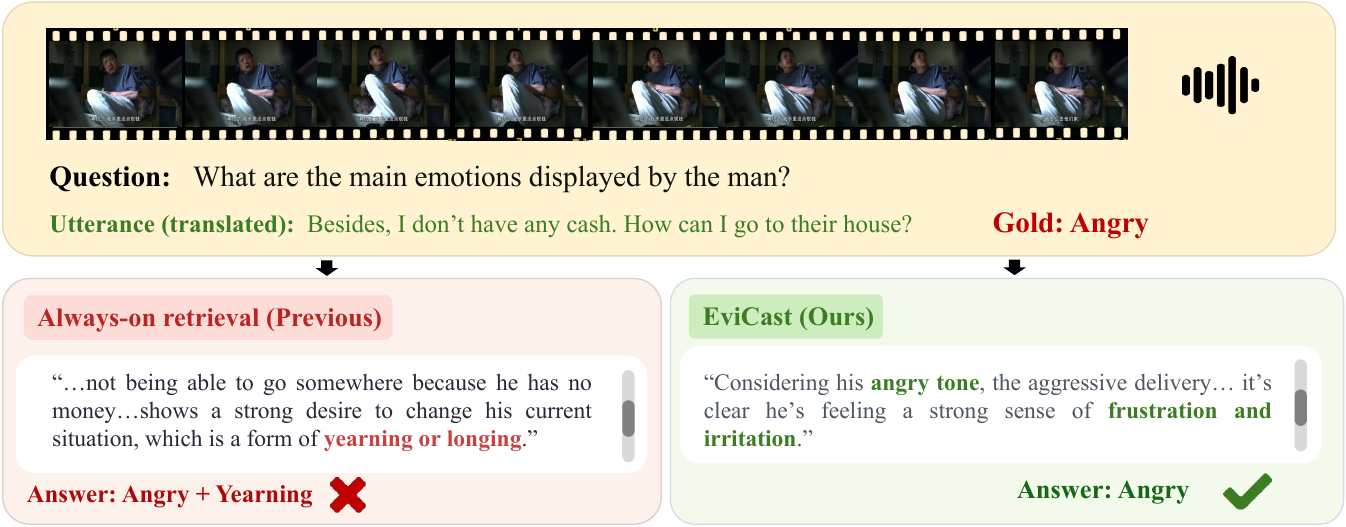}
  \caption{EMER example with video frames, recorded generator reasoning excerpts, and the EviCast configuration-level decision.}
  \label{fig:emer_qualitative_case}
\end{figure*}

Figure~\ref{fig:decision_plane} projects the 10 configurations onto the two-gate decision plane in Eq.~\eqref{eq:deployment_rule}. RAMER-AV has a slightly positive utility LCB but a separability margin below 5 points, so the separability gate prevents deployment based on this small estimate. RAMER-AL has a margin of 44.44 points but a utility LCB of $-4.34$ points, so the utility gate rejects a retrievable signal that does not yield actual gain. RAVDESS and HotpotQA satisfy both conditions and achieve large improvements on independent test sets. These cases show that evidence separability and model utility impose distinct deployment requirements whose conjunction defines an executable decision. Figure~\ref{fig:emer_qualitative_case} shows an EMER case in which the configuration-level decision retains the correct Base answer instead of the extra emotion label produced by retrieval.

\subsection{Reusable Calibration Across Retrievers}
\label{sec:retriever_reuse}

The candidate set fixes the HotpotQA questions, generator, and prompt. BGE-M3, BM25, TF-IDF, and RRF form the base group. Additional dense, sparse, and fusion variants rank passages using titles, lead sentences, or text truncated to 128 tokens. The set contains 12 deployable configurations and two controls: Oracle prioritizes the two annotated supporting passages, and Anti-BGE reverses the BGE-M3 ranking. All 14 configurations share one EviCast-B calibration. Appendix Table~S4 reports their forecasts, Direct estimates, independent test gains, absolute errors of both estimators, and deployment decisions.

For BGE-title, EviCast predicts a $31.21$-point F1 gain against an observed $31.05$-point gain, giving an absolute error of $0.16$ points. The errors remain $0.65$ and $2.47$ points when ranking uses lead sentences and 128-token inputs, respectively. The same response profile yields errors of $1.26$ points for BM25-lead and $4.12$ points for BGE--TF-IDF fusion, compared with Direct's $13.42$ and $14.64$ points.

Table~\ref{tab:retriever_reuse} evaluates the accuracy and efficiency of shared calibration. Across 14 HotpotQA retrieval configurations, one EviCast calibration uses 96 generation calls and makes 14 of 14 correct deployment decisions, with an MAE of 2.75 points and Pearson and Spearman correlations of 0.983 and 0.890. Relative to Direct, EviCast reduces forecast error by 71.3\% and generation calls by 80\%. The 96 calls comprise Base, Supportive, and Misleading responses for 32 calibration queries ($3\times32$). Direct evaluates Base and an Actual retrieval condition for each of the 14 configurations, requiring $15\times32=480$ calls.

\begin{table}[!tb]
  \centering
  \caption{Shared calibration on 14 HotpotQA configurations. MAE is in pp.}
  \label{tab:retriever_reuse}
  \footnotesize
  \setlength{\tabcolsep}{3pt}
  \fitwidth{\linewidth}{%
\begin{tabular}{@{}cccccc@{}}
      \toprule
      Method & Agree & Pearson & Spearman & MAE (pp) $\downarrow$ & Calls $\downarrow$ \\
      \midrule
      EviCast & \strongresult{14/14} & \strongresult{0.983} & \strongresult{0.890} & \strongresult{2.75} & \strongresult{96} \\
      Direct & 14/14 & 0.818 & 0.332 & 9.59 & 480 \\
      \bottomrule
    \end{tabular}}
\end{table}

\begin{figure*}[!tb]
  \centering
  \includegraphics[width=\linewidth]{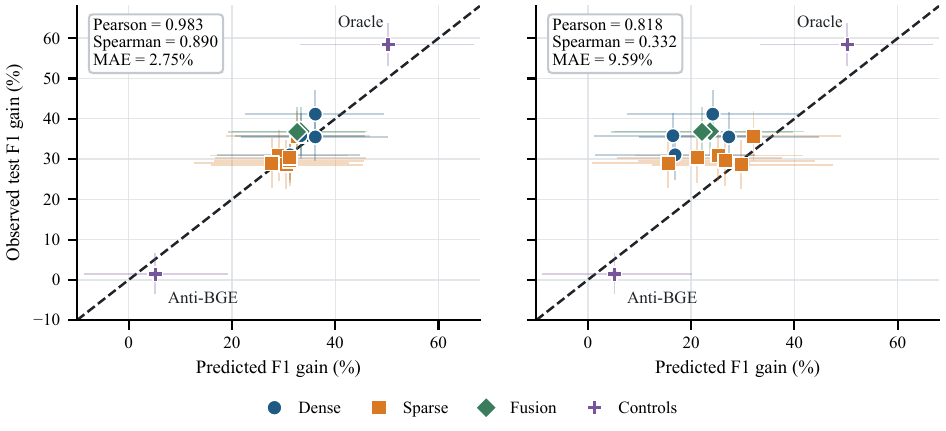}
  \caption{Forecast and independent test utility for 14 HotpotQA retrieval configurations, with EviCast on the left and Direct calibration on the right. Error bars show 95\% CIs.}
  \label{fig:utility_scatter}
\end{figure*}

Across 12 standard deployable configurations, EviCast obtains 12 of 12 correct decisions and an MAE of 2.22 points, compared with 10.20 points for Direct, while reducing generation calls by 76.9\%. These results support the decomposition in Fig.~\ref{fig:evicast}, where one generator response profile serves dense, sparse, and fusion retrievers through their state distributions. Figure~\ref{fig:utility_scatter} shows that this shared profile places every configuration close to the ideal diagonal, and Appendix Fig.~S1 shows how diagnostic savings scale with retriever count.

\subsection{Utility Forecasting Under Matched Budgets}
\label{sec:budget_comparison}

We compare EviCast, Direct, retrieval performance prediction (RPP), unnormalized Weighted Information Gain (U-WIG), the score-magnitude-and-variance predictor U-SMV, BERT-based Query Performance Prediction (BERT-QPP), and QualT5 on HotpotQA using Qwen2.5-3B-Instruct. The study fixes 128 calibration queries, 512 disjoint test queries, and BGE-M3, BM25, TF-IDF, and RRF. Fifty label-independent calibration permutations produce paired subsamples shared by every method and budget; all estimators and budgets are fixed before evaluation.

For answer-generation budget $C$, EviCast uses $\lfloor C/4\rfloor$ queries for Base and three topology states. Direct and comparison predictors use $\lfloor C/5\rfloor$ queries for the shared Base and four Actual conditions. EviCast recomputes stratum-conditioned responses and state frequencies on each subset, while Direct estimates the paired mean gain. We first average absolute forecast errors across the four retrievers and then average these MAEs across repetitions. The target is each retriever's mean F1 gain on the fixed test set.

U-WIG averages the two retrieved-passage scores, while U-SMV combines their magnitude and dispersion~\cite{tao2014query,dado2026predicting}. BERT-QPP supplies a supervised query-passage performance score~\cite{arabzadeh2021bertqpp}, and QualT5 uses the maximum learned passage-quality score among the same passages~\cite{chang2024neural}. All scalar baselines use the same calibration-only linear map from their native scores to F1 gain.

RPP adapts the 30-feature pre-generation utility predictor of Tian et al.~\cite{tian2026predicting} to the 10-candidate HotpotQA pool and fixed top-2 generator evidence. Features include retrieval-score statistics, embedding geometry and coherence, BERT-QPP, readability, QualT5, and teacher-forced generator context likelihood. Local Normalized Query Commitment (NQC) normalizes the top-two score deviation by the absolute mean candidate score, while coherence compares the top five and bottom five candidates. We use E5-small-v2, qt5-small, and the released mean average precision (MAP)-trained BERT-QPP checkpoint. Each retriever uses ordinary least squares with an intercept and calibration-only feature standardization. All 30 features enter the regression.

Budgets count answer generations. RPP additionally incurs teacher-forced likelihood scoring, embedding and quality-model inference, and feature computation, none of which EviCast requires. Appendix~I specifies the sampling seeds.

\begin{table}[!tb]
  \centering
  \caption{Matched answer-generation budgets on HotpotQA with Qwen2.5-3B-Instruct and four retrievers. MAE is in pp, and lower is better.}
  \label{tab:budget_comparison}
  \label{tab:appendix_budget_full}
  \footnotesize
  \setlength{\tabcolsep}{3pt}
  \fitwidth{\linewidth}{%
\begin{tabular}{@{}ccccc@{}}
    \toprule
    Method & 160 & 320 & 480 & 512 \\
    \midrule
    EviCast & \strongresult{5.73} & \strongresult{4.34} & 3.91 & 3.86 \\
    Direct & 6.85 & 4.89 & 3.94 & 3.92 \\
    RPP~\cite{tian2026predicting} & 15.25 & 5.81 & 4.88 & 4.77 \\
    U-WIG~\cite{dado2026predicting} & 7.04 & 4.89 & 3.95 & 3.94 \\
    U-SMV~\cite{tao2014query} & 6.99 & 4.83 & 3.84 & 3.79 \\
    BERT-QPP~\cite{arabzadeh2021bertqpp} & 6.99 & 4.86 & 3.89 & 3.87 \\
    QualT5~\cite{chang2024neural} & 6.77 & 4.68 & \strongresult{3.70} & \strongresult{3.70} \\
    \bottomrule
  \end{tabular}}
\end{table}

Table~\ref{tab:budget_comparison} shows that EviCast is most accurate in the low-budget regime that matters most for pre-deployment diagnosis. With 160 generations, EviCast attains the lowest MAE of 5.73 points, which is $15.4\%$ lower than QualT5, the strongest alternative, $16.4\%$ lower than Direct, and $62.4\%$ lower than RPP. With 320 generations, EviCast again attains the lowest MAE of 4.34 points, compared with 4.68 for QualT5, 4.89 for Direct, and 5.81 for RPP. Its advantage over Direct, RPP, and U-WIG holds at every budget, including the full budget of 512 generations, where EviCast uses all 128 calibration queries and the comparison predictors use 102 queries and 510 generations. Spending a fixed generation budget on shared response states therefore yields more accurate forecasts than spending it on per-retriever Actual outputs, and the return is largest when calibration budgets are tight.

EviCast is also the most stable predictor across calibration subsamples and attains the lowest 95th-percentile MAE of all seven methods at every budget, as reported in Appendix Table~S20.

\begin{table}[!tb]
  \centering
  \caption{Decision agreement and MAE (pp) across datasets and generators.}
  \label{tab:generalization}
  \footnotesize
  \setlength{\tabcolsep}{3pt}
  \fitwidth{\linewidth}{%
\begin{tabular}{@{}ccc@{}}
    \toprule
    Setting & Agree & MAE $\downarrow$ \\
    \midrule
    MuSiQue / Ours + Qwen~\cite{xu2025qwen25omni} & \strongresult{4/4} & \strongresult{1.53} \\
    MuSiQue / Ours + Phi-4~\cite{abouelenin2025phi4mini} & \strongresult{4/4} & \strongresult{1.55} \\
    HotpotQA / Ours + Qwen~\cite{xu2025qwen25omni} & \strongresult{4/4} & 4.59 \\
    HotpotQA / Ours + Phi-4~\cite{abouelenin2025phi4mini} & \strongresult{4/4} & \strongresult{3.58} \\
    2Wiki / Ours + Qwen~\cite{xu2025qwen25omni} & 3/4 & 4.70 \\
    2Wiki / Ours + Phi-4~\cite{abouelenin2025phi4mini} & \strongresult{4/4} & \strongresult{1.15} \\
    \midrule
    Three-dataset total & \strongresult{23/24} & \strongresult{2.85} \\
    Ours + Qwen2.5-Text~\cite{yang2024qwen25} (3B+7B) & \strongresult{8/8} & \strongresult{3.67} \\
    \bottomrule
  \end{tabular}}
\end{table}

\subsection{Generalization Across Datasets And Generators}
\label{sec:generalization}

\noindent\textbf{Cross-Dataset And Cross-Generator Validation.}\enspace Table~\ref{tab:generalization} evaluates topology-aware diagnosis on MuSiQue and HotpotQA with two generators and four retrievers. EviCast makes 16 of 16 decisions correctly with an overall MAE of 2.81 points. Each dataset--generator pair requires 512 shared calibration generations rather than Direct's 640. Complete per-configuration forecasts and independent test gains are provided in Appendix~D.

\noindent\textbf{Prospective Transfer To A New Dataset.}\enspace We prospectively transfer the topology-aware diagnosis to 2Wiki under a protocol frozen before producing retrieval rankings or generator outputs. The study uses 128 shared calibration examples and 512 disjoint test examples. It evaluates two generators with BGE-M3, BM25, TF-IDF, and RRF under fixed top-2 retrieval. Across eight configurations, EviCast makes 7/8 correct decisions with no false enables. Forecast gains achieve Pearson and Spearman correlations of 0.955 and 0.905 with independent test gains. Direct makes 6/8 correct decisions and requires 640 rather than 512 generation calls per generator. Appendix Tables~S10 and~S11 report the aggregate and per-configuration results. Across MuSiQue, HotpotQA, and 2Wiki, EviCast makes 23 of 24 frozen decisions correctly with an aggregate MAE of 2.85 points.

\noindent\textbf{Transfer To Text-Only Generators Across Model Scales.}\enspace We evaluate frozen topology-aware diagnosis on standard text-only generators at different scales. The study uses Qwen2.5-3B-Instruct and Qwen2.5-7B-Instruct on HotpotQA with BGE-M3, BM25, TF-IDF, and RRF. For each generator, EviCast shares 128 calibration examples and fixes the top-2 evidence states, anchor, separability threshold, and deployment rule. Each retriever is evaluated on 512 disjoint test queries. Table~\ref{tab:text_scale} reports the results. Across both scales, EviCast makes all eight deployment decisions correctly with no false enables or rejects. Its aggregate MAE is 3.67 points, compared with 6.28 for EviCast-B and 3.79 for Count. These results extend the shared intervention and evidence-state decomposition to text-only generators at two parameter scales.

\begin{table}[!tb]
\centering
  \setlength{\tabcolsep}{2pt}
  \caption{Frozen transfer to text-only generators at the 3B and 7B parameter scales.}
  \label{tab:text_scale}
  \footnotesize
  \fitwidth{\linewidth}{%
\begin{tabular}{cccccc}
    \toprule
    Generator & Params. & Configs. & Agree & \shortstack{False\\enable} & \shortstack{MAE\\(pp)} $\downarrow$ \\
    \midrule
    Qwen2.5-Instruct & 3B & 4 & \strongresult{4/4} & \strongresult{0} & 3.86 \\
    Qwen2.5-Instruct & 7B & 4 & \strongresult{4/4} & \strongresult{0} & \strongresult{3.49} \\
    \midrule
    Overall & 3B and 7B & 8 & \strongresult{8/8} & \strongresult{0} & \strongresult{3.67} \\
    \bottomrule
  \end{tabular}}
\end{table}

\subsection{Ablation And Mechanism Analysis}
\label{sec:ablation}

\noindent\textbf{Two-Gate Decision Rule.}\enspace Table~\ref{tab:appendix_gate_ablation} removes either gate across the same 10 configurations. Separability alone falsely enables RAER-EMER, RAMER-AL, and RAMER-VL. Utility alone falsely enables RAMER-AV and misses RAMER-L. The complete rule eliminates false enables while retaining both strongly beneficial configurations. The main experiments fix the separability threshold at $\tau=0.05$. Holding the utility gate $L_j>0$ fixed, a sweep over the same 10 configurations retains 9/10 agreement with no false enables throughout $0.05\leq\tau\leq0.30$. Appendix Table~S15 reports the complete sweep.

\begin{table}[!tb]
  \centering
  \caption{Two-gate ablation on 10 task configurations. Retrieval is enabled only when all included gates pass. The final row is EviCast.}
  \label{tab:appendix_gate_ablation}
  \small
  \setlength{\tabcolsep}{5pt}
  \begin{tabular}{ccccc}
    \toprule
    \multicolumn{3}{c}{Deployment decisions} & \multicolumn{2}{c}{Gates included} \\
    \cmidrule(lr){1-3}\cmidrule(lr){4-5}
    \shortstack{Correct\\decisions $\uparrow$} & \shortstack{False\\enables $\downarrow$} & \shortstack{Missed\\benefits $\downarrow$} & \shortstack{Relevant\\evidence} & \shortstack{Positive\\gain} \\
    \midrule
    7/10 & 3 & 0 & $\checkmark$ & --- \\
    8/10 & 1 & 1 & --- & $\checkmark$ \\
    \strongresult{9/10} & \strongresult{0} & 1 & \strongresult{$\checkmark$} & \strongresult{$\checkmark$} \\
    \bottomrule
  \end{tabular}
\end{table}

\begin{table*}[!tb]
  \centering
  \setlength{\tabcolsep}{4pt}
  \footnotesize
  \caption{Decision agreement and MAE (pp) along the state refinement of Proposition~\ref{prop:refinement}, with Direct calibration as reference.}
  \label{tab:state_ablation}
  \fitwidth{\linewidth}{
  \begin{tabular}{cccccccc}
    \toprule
    Method & MuSiQue agree & MuSiQue MAE & HotpotQA agree & HotpotQA MAE & Overall agree & Overall MAE & Gen. calls \\
    \midrule
    EviCast-B & 8/8 & 7.24 & 8/8 & 9.35 & 16/16 & 8.30 & 384 \\
    Count & 8/8 & 3.49 & 8/8 & 4.25 & 16/16 & 3.87 & 512 \\
    \strongresult{EviCast-T} & \strongresult{8/8} & \strongresult{1.54} & \strongresult{8/8} & \strongresult{4.08} & \strongresult{16/16} & \strongresult{2.81} & \strongresult{512} \\
    \midrule
    \multicolumn{8}{c}{\textbf{\textit{Direct calibration reference}}} \\
    \midrule
    Direct & 8/8 & 1.79 & 8/8 & 3.75 & 16/16 & 2.77 & 640 \\
    \bottomrule
  \end{tabular}}
\end{table*}

\noindent\textbf{Evidence-State Representation.}\enspace Table~\ref{tab:state_ablation} traces the refinement predicted by Proposition~\ref{prop:refinement} while fixing datasets, retrievers, generators, and retrieval statistics. EviCast-B interpolates between the supportive and misleading endpoints and yields an overall MAE of 8.30 points. Count separates zero, one, and two correct passages and reduces the MAE to 3.87 points. EviCast-T further separates terminal and bridge roles and reduces the MAE to 2.81 points, lowering forecast error by 66.1\% and 27.4\%, respectively, while preserving 16 of 16 correct deployment decisions. EviCast-T matches the 2.77-point MAE of per-retriever Direct calibration within 0.04 points while requiring 20\% fewer diagnostic calls, so its accuracy comes from structured, reusable response calibration rather than a larger generation budget.

\begin{figure}[!tb]
  \centering
  \includegraphics[width=\linewidth]{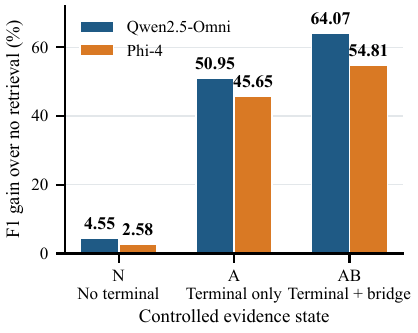}
  \caption{Evidence-role effects on two-hop MuSiQue questions. Bars show F1 gains over no retrieval for Qwen and Phi-4, and every intervention supplies two passages.}
  \label{fig:topology_response_plot}
\end{figure}

\noindent\textbf{Why Evidence Topology Matters.}\enspace Figure~\ref{fig:topology_response_plot} and Appendix Table~S6 show why correct-passage counts are insufficient. States $N$ and $A$ each contain one correct supporting passage, but $N$ provides only bridge information, whereas $A$ directly contains terminal answer information. Their F1 gains are 4.55 and 50.95 points for Qwen and 2.58 and 45.65 points for Phi-4. Thus, one correct passage does not define a utility-homogeneous state. Reusable profiles must distinguish terminal and bridge roles.

This contrast is exactly the refinement bias of Proposition~\ref{prop:refinement}. Because Count and EviCast-T use the same number of diagnostic calls, their accuracy difference in Table~\ref{tab:state_ablation} isolates the value of distinguishing evidence roles. Appendix Tables~S6 and~S7 report the responses for every stratum and the state distribution of every retriever.

\subsection{Empirical Verification Of Theoretical Properties}
\label{sec:theory_check}

\begin{table}[!tb]
  \centering
  \caption{Empirical verification of the theoretical properties on MuSiQue and HotpotQA. Deviations are in pp relative to a finer-state reference.}
  \label{tab:theory_check}
  \footnotesize
  \setlength{\tabcolsep}{3pt}
  \fitwidth{\linewidth}{%
  \begin{tabular}{@{}cccc@{}}
    \toprule
    Property & Statistic & MuSiQue & HotpotQA \\
    \midrule
    Refinement (Prop.~\ref{prop:refinement}) & \shortstack{Deviation of EviCast-B /\\Count / EviCast-T} & 7.55 / 3.80 / \strongresult{0.31} & 5.95 / 1.58 / \strongresult{0.27} \\
    Sufficiency (Assump.~\ref{ass:sufficiency}) & Smallest $p$ & \strongresult{0.38} & \strongresult{0.26} \\
    Efficiency (Prop.~S2) & Largest $J^\star$ & \strongresult{2.15} & \strongresult{2.09} \\
    \bottomrule
  \end{tabular}}
\end{table}

\noindent\textbf{Refinement.}\enspace Table~\ref{tab:theory_check} confirms each property. Relative to a reference that additionally separates retrieved sets without any supporting passage from bridge-only sets, EviCast-T deviates by only 0.31 and 0.27 F1 points on MuSiQue and HotpotQA, compared with 7.55 and 5.95 points for EviCast-B. The topology-aware map is therefore already sufficient, and its remaining forecast error reflects calibration sampling rather than state representation. Corollary~\ref{cor:binary} also explains when EviCast-B suffices. Post hoc, the test-set linearity gap is $-0.31$ points on the 14-configuration reuse split, where EviCast-B is highly accurate, but $-15.30$ and $-10.00$ points for Qwen and Phi-4 on the topology-aware HotpotQA split.

\noindent\textbf{Sufficiency.}\enspace Stratified permutation tests detect no retriever dependence of the response within any state in all 15 tests, including the 12-retriever reuse study (all $p\ge0.26$).

\noindent\textbf{Efficiency.}\enspace The influence-function ratio $\nu_j/\sigma_j$ lies between 0.73 and 0.89, and $J^\star_j\le2.15$ in all 16 configurations, so every candidate set in our experiments lies in the regime where shared calibration is the more efficient estimator. Appendix~\mbox{E-E} reports the complete statistics.

\section{Conclusion}
\label{sec:conclusion}

Retrieval deployment should depend on how a target generator uses evidence, not only on its availability. EviCast forecasts utility by combining shared interventional response calibration with retriever-specific evidence-state distributions. A state-refinement analysis characterizes when its binary and topology-aware instantiations are exact. Experiments show that response calibration transfers across retrievers and that terminal and bridge roles explain utility differences hidden by passage counts. Across three datasets, EviCast decides 23 of 24 topology configurations correctly, and in the low-budget regime it forecasts utility more accurately than recent performance predictors. On HotpotQA, EviCast makes all 14 deployment decisions correctly while reducing diagnostic calls by 80\%. These results shift RAG from default retrieval toward utility-guided deployment, with one calibration shared by multiple retrievers. Future work will extend evidence states to variable depths, iterative retrieval, and multimodal knowledge environments.

\section*{Acknowledgments}
This work was supported in part by the National Natural Science Foundation of China under Grants 62606208, 62606209, 62227807, and U24B20186, by the National Science and Technology Major Project on Brain Science and Brain-like Intelligence Technology under Grants 2021ZD0200600 and 2021ZD0200408, and by the Talent Scientific Fund of Lanzhou University.
We also acknowledge support from the WQ \& UCAS Research Academy Intelligent Computing Center (WRA-ICC) and the Supercomputing Center of Lanzhou University.

\bibliographystyle{IEEEtran}
\bibliography{references}
\makeatletter
\def\@IEEEBIOphotowidth{1in}
\def\@IEEEBIOphotodepth{1.25in}
\def\@IEEEBIOhangwidth{1.1in}
\def\@IEEEBIOhangdepth{1.25in}
\def\@IEEEBIOskipN{1\baselineskip}
\makeatother
\section*{Author Biographies}

\begin{IEEEbiography}[{\includegraphics[width=1in,height=1.25in,clip,keepaspectratio]{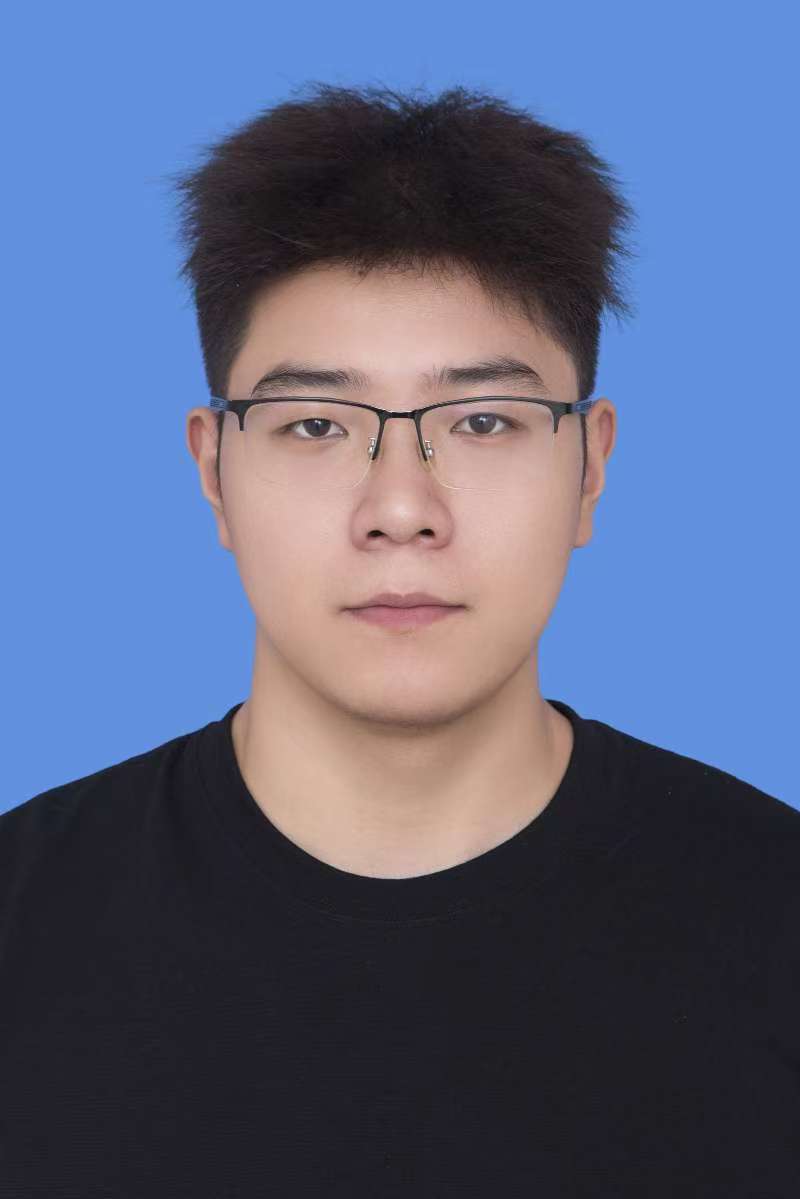}}]{Yibo Wang}
is currently pursuing the M.S. degree at the Gansu Provincial Key Laboratory of Wearable Computing, School of Information Science and Engineering, Lanzhou University, Lanzhou, China. He received the B.S. degree from Dalian University of Technology, Dalian, China. His research interests include multimodal large language models, affective computing, and reinforcement learning.
\end{IEEEbiography}

\begin{IEEEbiography}[{\includegraphics[width=1in,height=1.25in,clip,keepaspectratio]{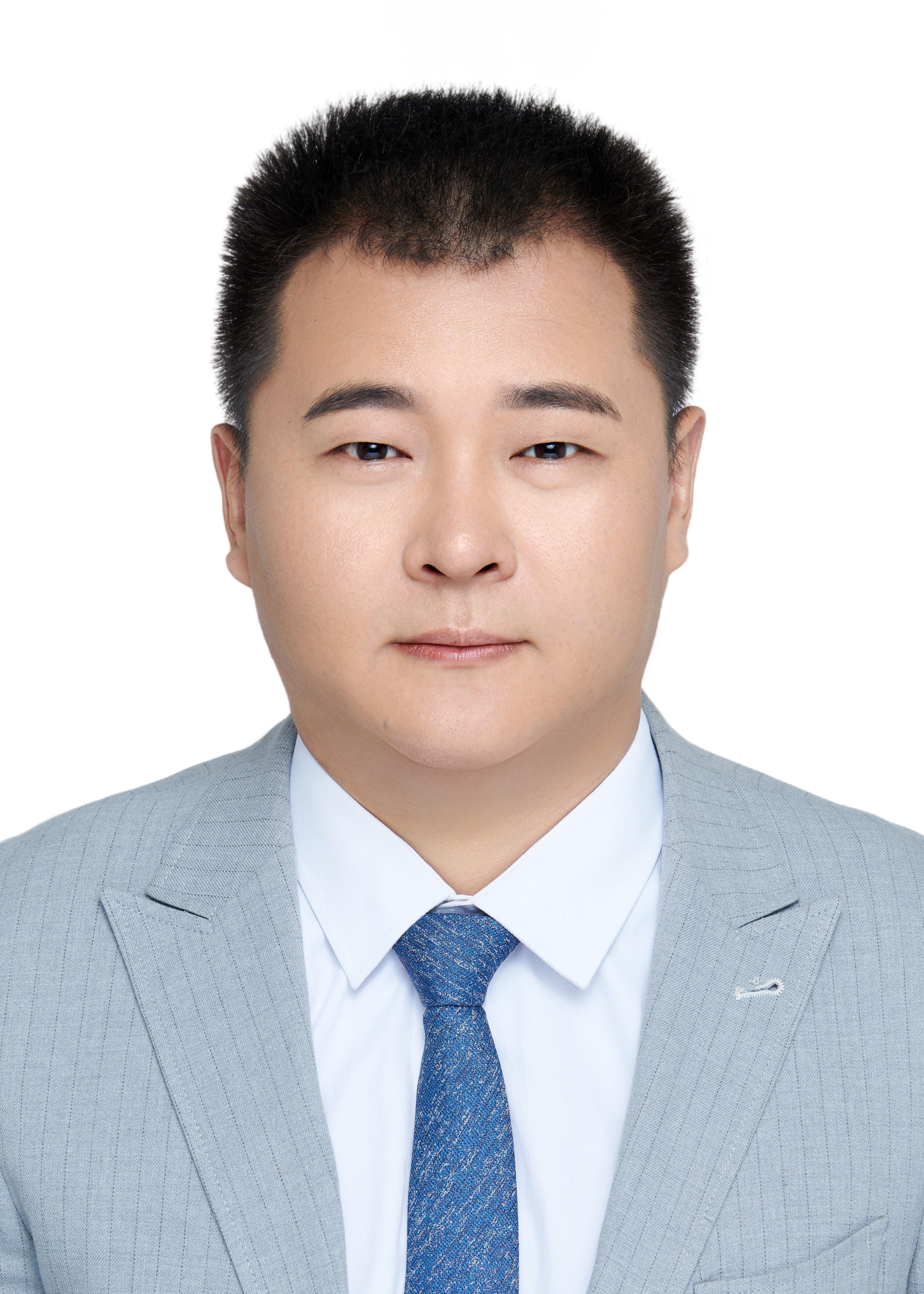}}]{Jisheng Dang}
received the Ph.D. degree from Sun Yat-sen University, China, advised by Prof. Jianhuang Lai and Prof. Huicheng Zheng. He worked as a research fellow at the NExT++ laboratory of the National University of Singapore, advised by Prof. Tat-Seng Chua. He is now a tenured associate professor at the School of Information Science and Engineering, Lanzhou University. His research interests include multimodal learning, video understanding, and embodied intelligence. He has published several papers as the first author or corresponding author in major journals and conferences including IEEE TPAMI/TIP/TNNLS/TITS/IJCAI/AAAI/ICLR/NeurIPS/PR.
\end{IEEEbiography}

\begin{IEEEbiography}[{\includegraphics[width=1.0in,height=1.25in,clip,keepaspectratio]{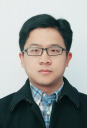}}]
{Meng Wang (Fellow, IEEE)}  received the B.E. and Ph.D. degrees (in the Special Class for the Gifted Young) from the Department of Electronic Engineering and Information Science, University of Science and Technology of China, Hefei, China, respectively. He is a Professor at the Hefei University of Technology, Hefei. His current research interests include multimedia content analysis, search, mining, recommendation, and large-scale computing. Dr. Wang was the recipient of the Best Paper Awards successively from the 17th and 18th ACM International Conference on Multimedia, the Best Paper Award from the 16th International Multimedia Modeling Conference, Best Paper Award from the 4th International Conference on Internet Multimedia Computing and Service, and Best Demo Award from the 20th ACM International Conference on Multimedia. 
\end{IEEEbiography}

\begin{IEEEbiography}[{\includegraphics[width=1in,height=1.25in,clip,keepaspectratio]{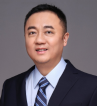}}]{Qi Tian (Fellow, IEEE)}
received the Ph.D. degree in ECE from the University of Illinois at Urbana-Champaign, Champaign, IL, USA, in 2002. He is currently the Chief Scientist of Huawei Terminal BG. He was the Chief Scientist in computer vision with Huawei Noah's Ark Laboratory from 2018 to 2020. Before he joined Huawei, he was a Full Professor with the Department of Computer Science, The University of Texas at San Antonio, San Antonio, TX, USA, from 2002 to 2019. He was listed in the top ten of the 2016 Most Influential Scholars in Multimedia by Aminer.org. He is an IEEE Fellow (Class 2016) and an Academician of the International Eurasian Academy of Sciences (elected in 2021), with 117,426 citations on Google Scholar.
\end{IEEEbiography}

\begin{IEEEbiography}[{\includegraphics[width=1in,height=1.25in,clip,keepaspectratio]{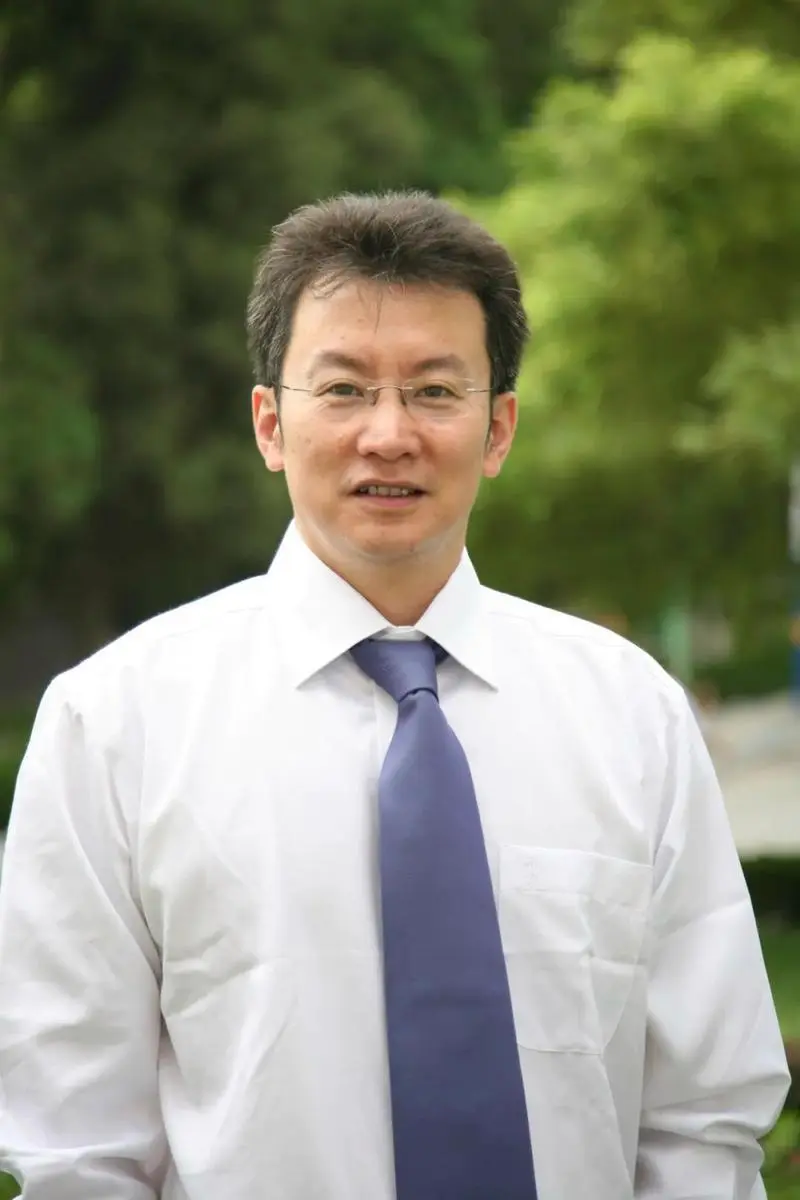}}]{Bin Hu (Fellow, IEEE)}
received the Ph.D. degree in computer science from the Institute of Computing Technology, Chinese Academy of Sciences, China, in 1998. Since 2008, he has been a Professor and Dean of the School of Information Science and Engineering, Lanzhou University. He has also held a guest professorship at ETH Zurich. He serves as Editor-in-Chief of IEEE Transactions on Computational Social Systems and is a Fellow of IET and AAIA. His research interests include pervasive computing, computational psychophysiology, data modeling, and artificial intelligence.
\end{IEEEbiography}

\begin{IEEEbiography}[{\includegraphics[width=1in,height=1.25in,clip,keepaspectratio]{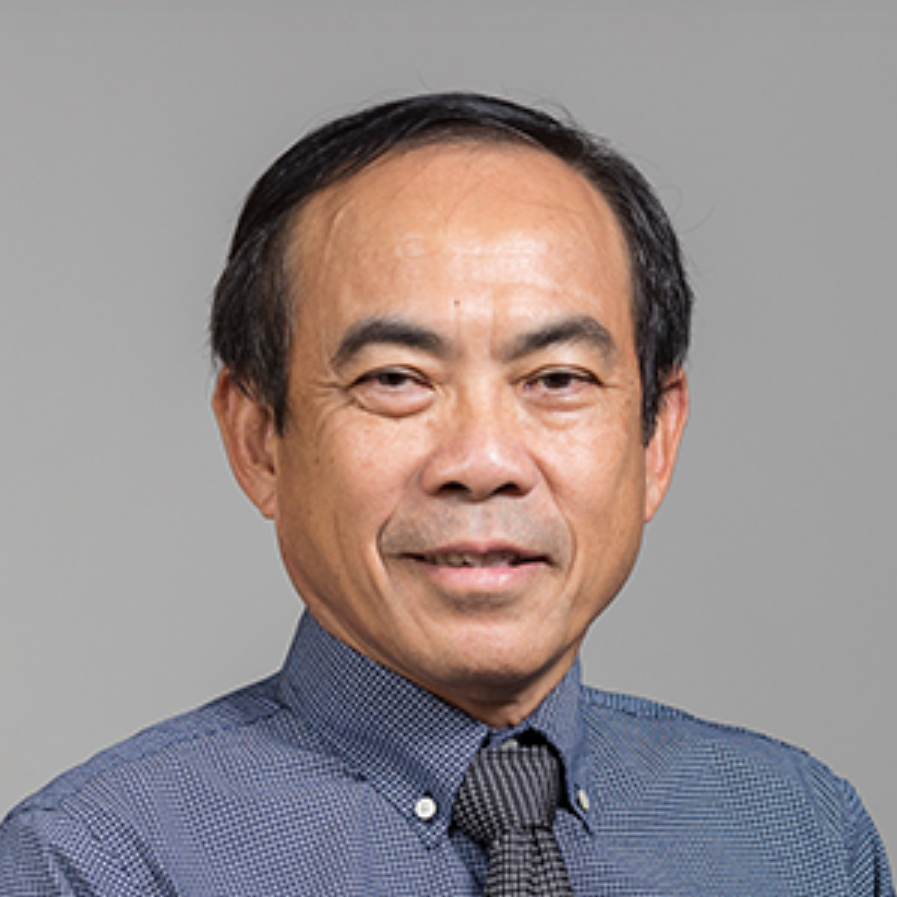}}]{Tat-Seng Chua}
received the Ph.D. degree from the University of Leeds, U.K. He is the KITHCT chair professor with the School of Computing, National University of Singapore, where he was the acting and founding dean of the School from 1998 to 2000. He is the co-director of NExT, a joint center between NUS and Tsinghua University, to develop technologies for live social media search. He is the 2015 winner of the prestigious ACM SIGMM Award. He is the chair of the Steering Committee of the ACM International Conference on Multimedia Retrieval (ICMR) and the Multimedia Modeling (MMM) conference series. He is also the general co-chair of ACM Multimedia 2005, ACM CIVR (now ACM ICMR) 2005, ACM SIGIR 2008, and ACM Web Science 2015. He serves on the editorial boards of four international journals. He is the co-founder of two technology startups in Singapore and a Fellow of the Singapore Academy of Sciences, with 114,163 citations on Google Scholar.
\end{IEEEbiography}

\vfill
\end{document}